\documentclass[letterpaper, 10 pt, conference]{ieeeconf}

\IEEEoverridecommandlockouts
\usepackage{subfig}

\usepackage{amsmath}
\usepackage{amssymb}
\usepackage{graphicx}
\usepackage{booktabs}

\usepackage{xcolor}
\usepackage{cite}
\usepackage[hidelinks]{hyperref}

\graphicspath{{figures/v2/}{figures/}}

\newcommand{\xs}{s}
\newcommand{\obs}{o}
\newcommand{\est}{\hat{s}}
\newcommand{\phis}{\phi}
\newcommand{\teacher}{\pi^{*}}
\newcommand{\act}{a^{*}}
\newcommand{\actp}{\hat{a}}
\newcommand{\Ls}{L_s}
\newcommand{\La}{L_a}
\newcommand{\ws}{w_s}
\newcommand{\wa}{w_a}

\newlength{\drtile}
\newlength{\drgap}
\newlength{\drsep}
\newlength{\drlab}
\newcommand{\drlabel}[1]{\makebox[\drlab][c]{\raisebox{\dimexpr0.5\drtile-2.5pt\relax}{\rotatebox[origin=c]{90}{\scriptsize #1}}}}
\newcommand{\drhead}[2]{\makebox[#1][c]{\scriptsize #2}}

\title{\LARGE \bf
Estimate, Don't Imitate: Reusing Differentiable State-Based Policies for Visuomotor Control
}

\newif\ifanonymous    \anonymousfalse
\newif\ifwithappendix \withappendixtrue

\ifanonymous
\author{Anonymous Authors}
\else
\author{%
Denis Shcherba$^{1,2,*}$,
Adrian Abel$^{1,2}$,
Eckart Cobo-Briesewitz$^{2}$,
Paul Mattes$^{1,2}$,\\
Wojciech Samek$^{1,2,4}$ and
Marc Toussaint$^{2,3}$%
\thanks{$^{1}$Fraunhofer Heinrich-Hertz-Institut, Berlin, Germany.}%
\thanks{$^{2}$Technische Universit\"at Berlin, Germany.}%
\thanks{$^{3}$Robotics Institute Germany.}%
\thanks{$^{4}$BIFOLD -- Berlin Institute for the Foundations of Learning and Data, Berlin, Germany.}%
\thanks{$^{*}$Corresponding author: \texttt{denis.shcherba@hhi.fraunhofer.de}}%
\thanks{Supplementary video: \url{https://youtu.be/G3_gS-dAztk}}%
}
\fi

\makeatletter
\IEEEaftertitletext{%
  \begin{minipage}{\textwidth}
    \centering
    \includegraphics[width=0.95\textwidth]{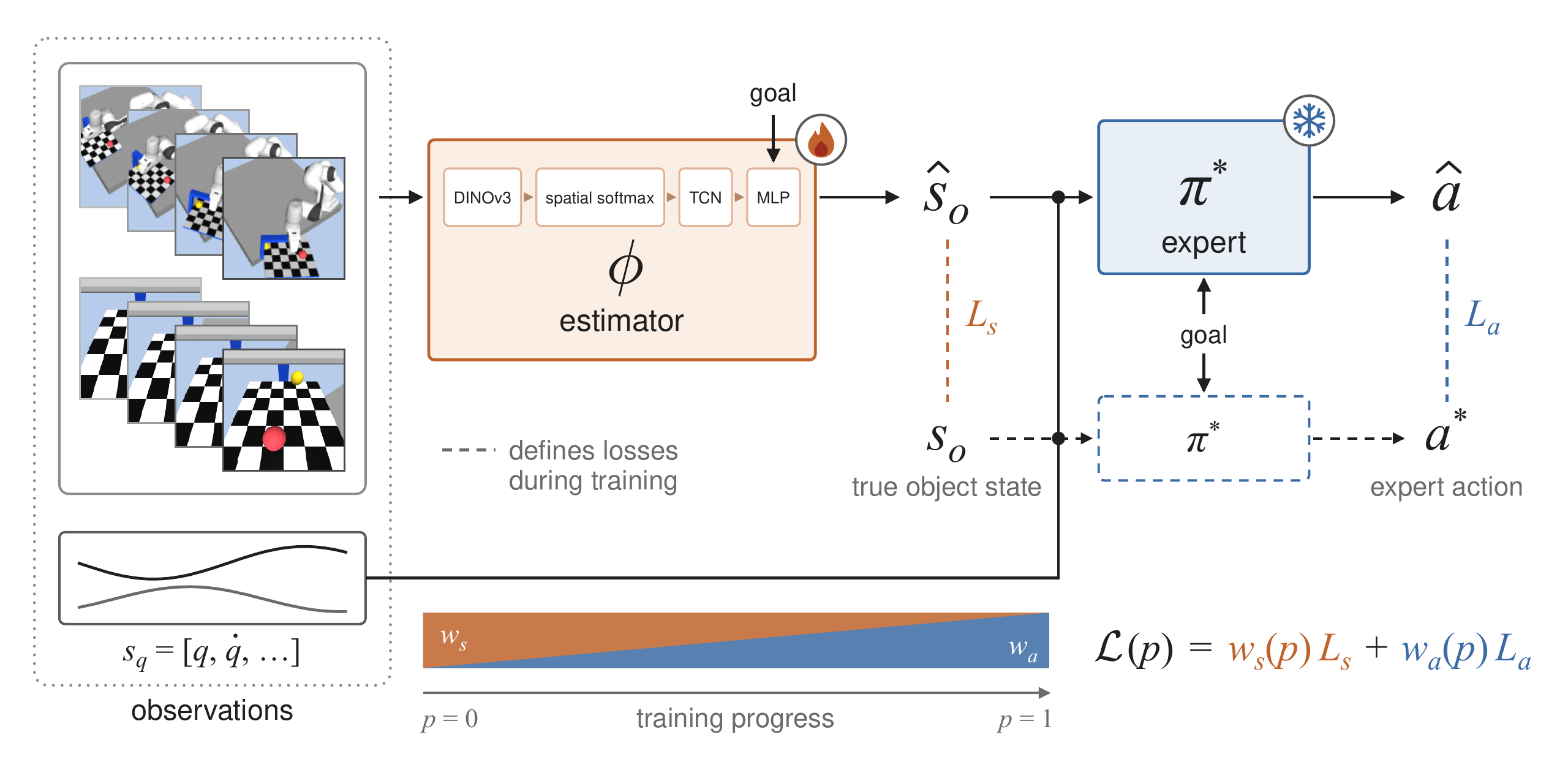}
    \def\@captype{figure}%
    \caption{Vision-based state estimation with a reused expert. The estimator $\phis$ maps images and the goal to $\hat{s}_o$, the object-state and goal-relative entries, composed with the proprioceptive state $s_q$ into the frozen expert's input $\est$. $\Ls$ compares $\hat{s}_o$ with $s_o$; $\La$ propagates gradients through the expert. Their weights follow a schedule in $p$. Only $\phis$ is updated.}
    \label{fig:method}
    \vspace{\baselineskip}
  \end{minipage}%
}
\makeatother

\begin{document}
\bstctlcite{IEEEexample:BSTcontrol}
\maketitle
\thispagestyle{empty}
\pagestyle{empty}

\begin{abstract}

Simulation-trained manipulation policies can exploit privileged state information to learn effective contact-rich behaviours, but deployment requires acting from partial observations such as noisy camera images. A common solution is teacher–student distillation, in which a visuomotor policy is trained to reproduce the actions of the privileged expert. This requires the student to jointly infer the task-relevant state and relearn the expert's action mapping that is already available. An alternative is to reuse the state-based expert and learn only a perceptual interface that reconstructs its missing state inputs. However, minimising the state estimate error alone does not necessarily minimise the downstream control error induced by these estimates. To bridge this gap, we train a visual state estimator using both direct state supervision and an action-consistency loss backpropagated through the frozen, differentiable expert. A scheduled objective first establishes a physically meaningful state estimate and progressively emphasises errors that affect the expert’s actions. Across five goal-conditioned manipulation tasks, retaining the expert consistently outperforms direct pixel-to-action imitation from the same expert demonstration corpus. We further demonstrate sim-to-real transfer on a physical Panda robot, achieving 76\% success without retraining the underlying expert.

\end{abstract}

\begin{figure*}[t]
    \centering
    \subfloat[PandaSphere]{\includegraphics[width=0.2\textwidth]{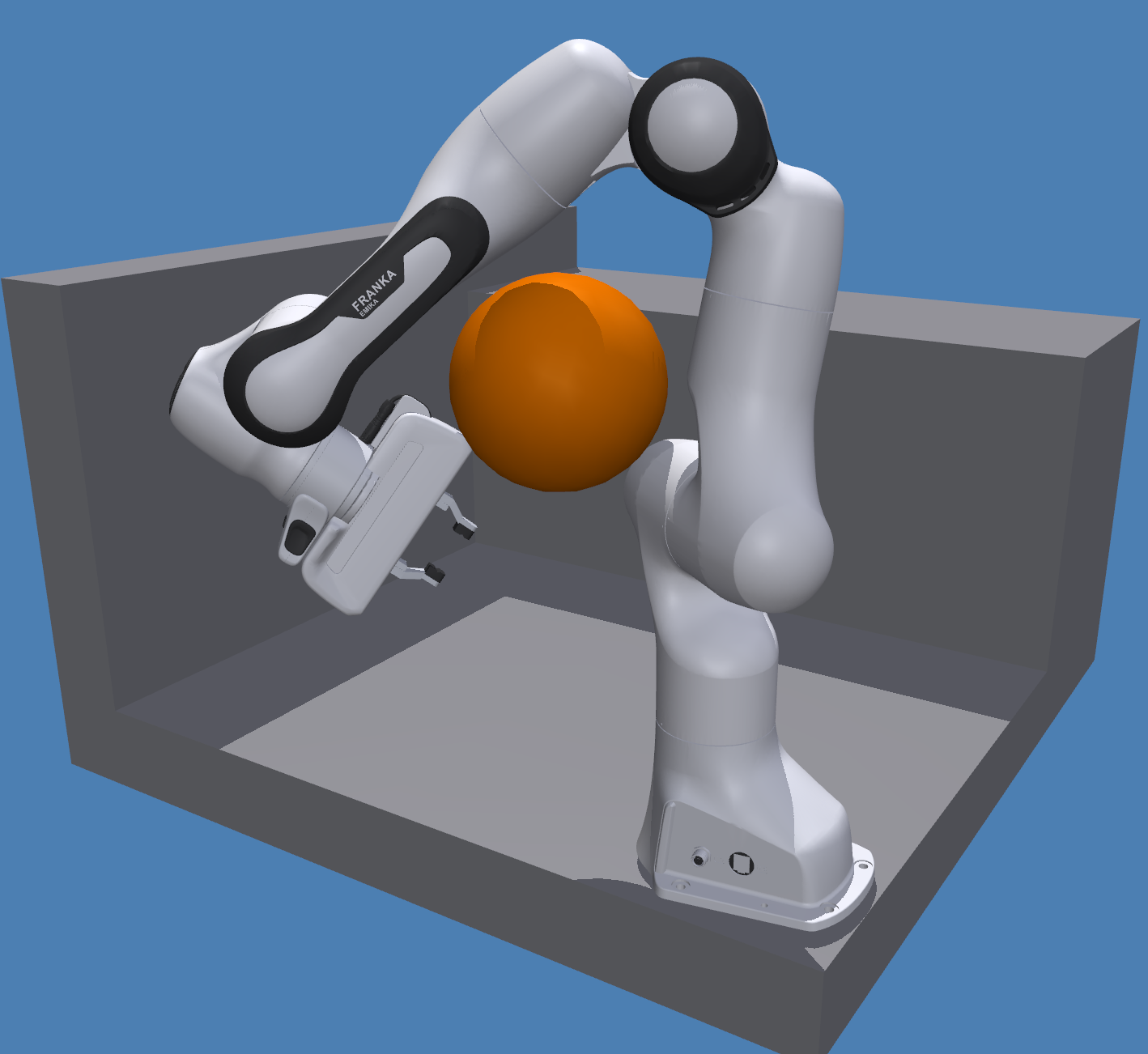}} \hfill
    \subfloat[PandaCube]{\includegraphics[width=0.2\textwidth]{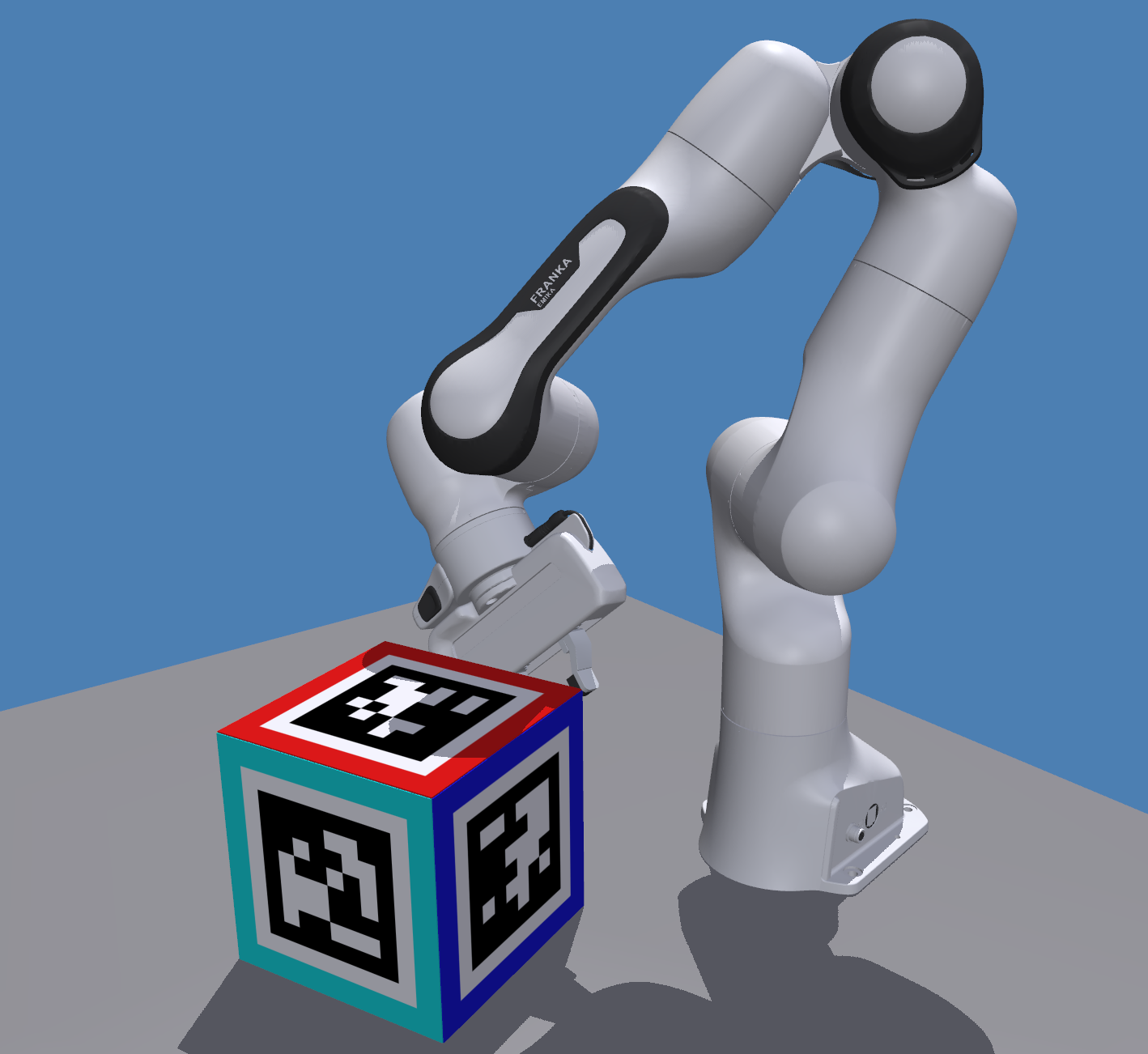}} \hfill
    \subfloat[TraySpiral]{\includegraphics[width=0.2\textwidth]{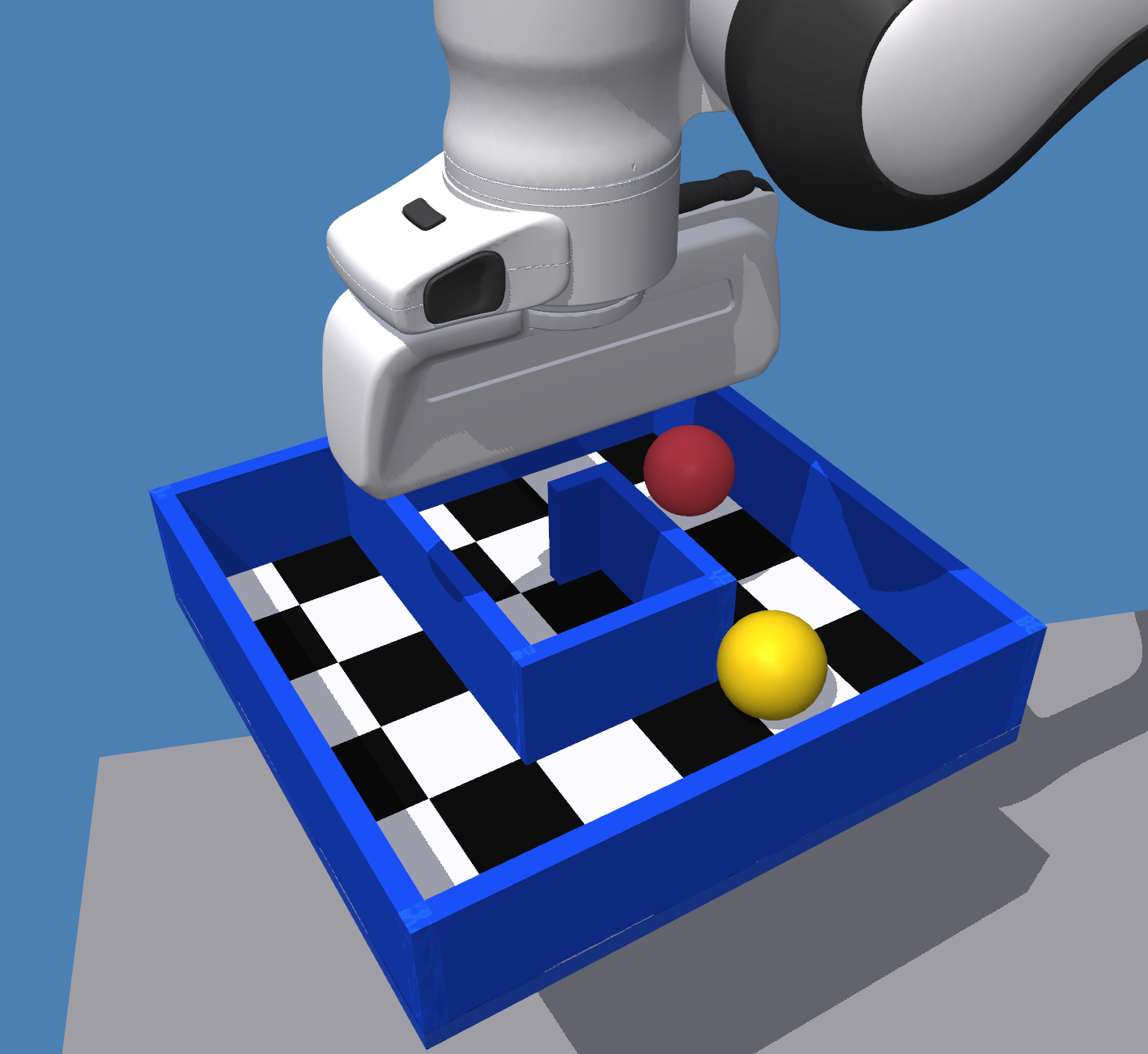}} \hfill
    \subfloat[TrayPlate]{\includegraphics[width=0.2\textwidth]{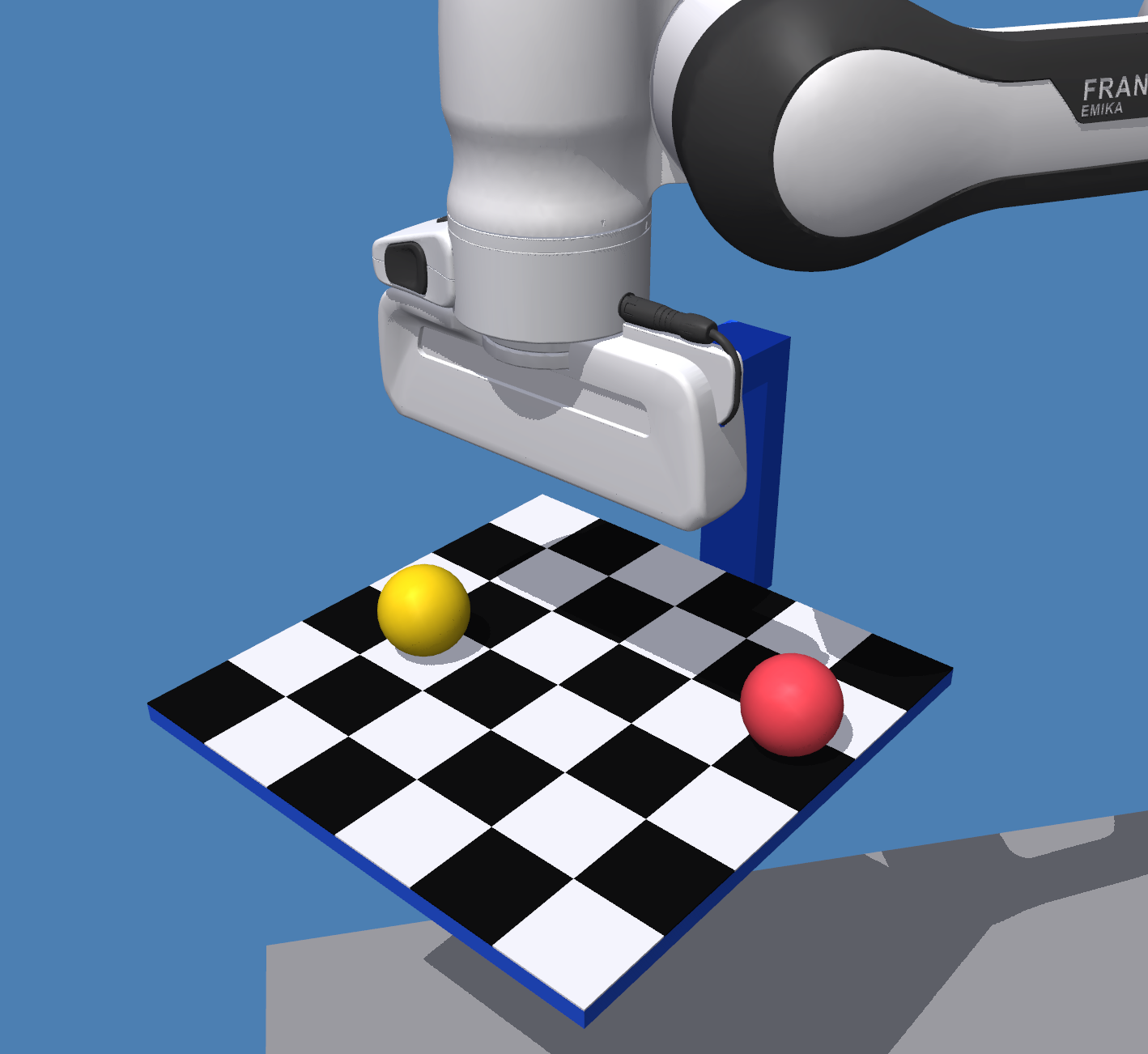}} \hfill
    \subfloat[AllegroCube]{\includegraphics[width=0.2\textwidth]{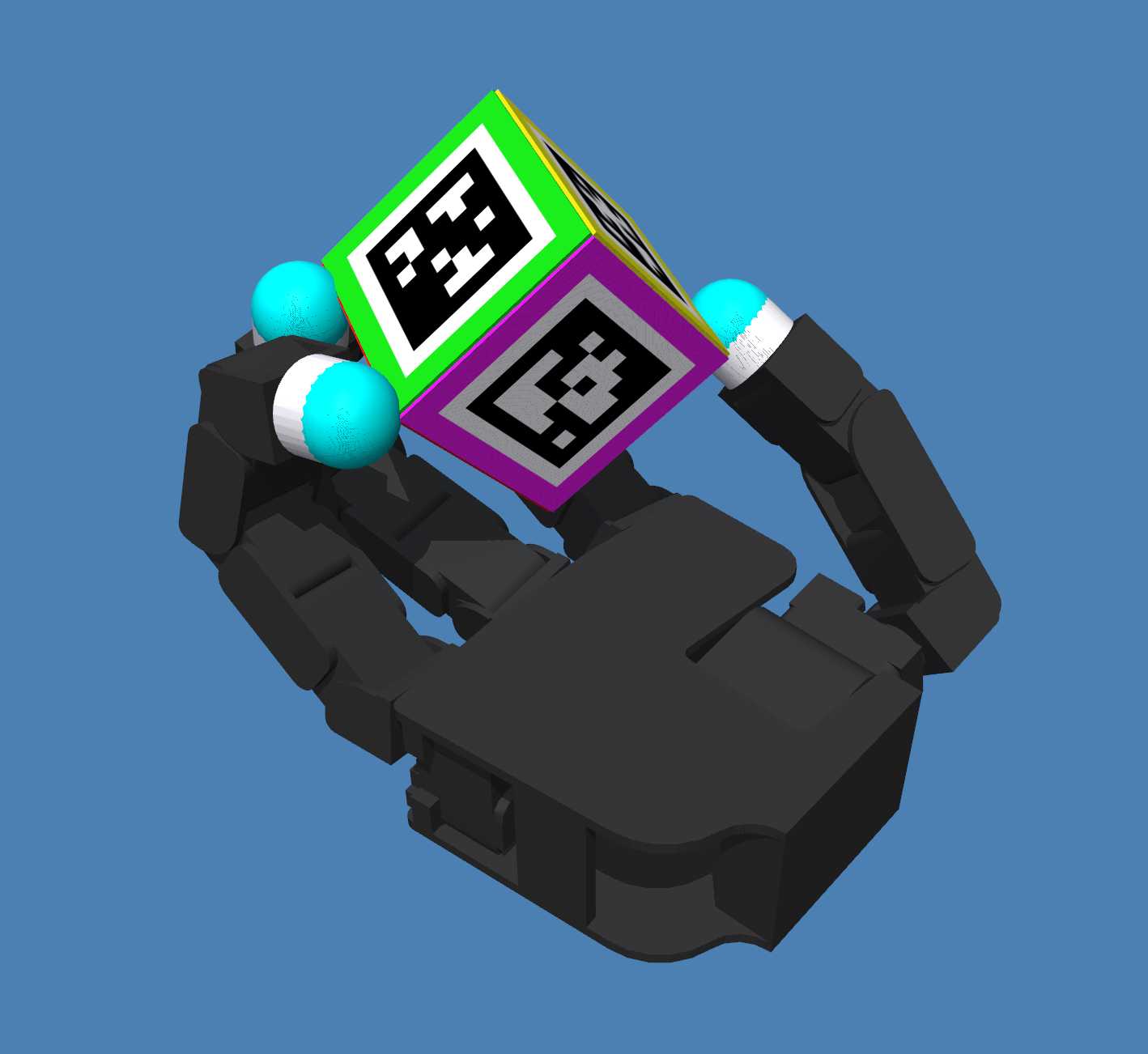}}
    \caption{The five environments: contact-rich manipulation tasks of varying difficulty, each with a single expert conditioned on a target object state.}
    \label{fig:environments}
\end{figure*}

\section{Introduction}

Advances in Reinforcement Learning (RL)~\cite{fujimoto2023td7, andrychowicz2017her} have led to strong performance in contact-rich robotic manipulation~\cite{toussaint2026combinedconstrainedsamplingreinforcement}. However, many of these methods rely on privileged state knowledge of the environment in order to converge to a valid policy, making them hard to transfer to the real world. Policies trained on partial observations of the environment, such as images, instead suffer from distribution shift when transferred from simulation to the real world. One popular approach to this shift is domain randomisation~\cite{tobin2017domainrandomizationtransferringdeep}, where a policy is trained on many instances of the same environment with different simulation parameters, at the cost of training performance, as the policy might struggle with finding good representations for the perceived observations. One prominent approach to overcoming this issue is the teacher--student formulation~\cite{chen2020lbc, lee2020locomotion}, where a teacher policy is first trained on full state knowledge of the system and a student policy conditioned on partial observations is subsequently trained to imitate the teacher. Related methods have also successfully trained visuomotor policies from data generated by trajectory-centric controllers, motion planners, and learned privileged experts~\cite{levine2016endtoendtrainingdeepvisuomotor, zhu2025learncontactrichmanipulationpolicies}. Yet, these approaches ultimately replace the controller that generated the demonstrations with a newly learned control policy, requiring the expert's control mapping to be recovered from scratch.

This raises a central question: \emph{if the control policy (teacher) is already available, is it necessary to relearn an observation-to-action policy (student) from scratch, or is it sufficient to learn how observations map into the expert's state inputs and retain the existing controller for downstream execution?}

 We refer to this formulation as \emph{sufficient state prediction for frozen-expert execution}. The learned perception module maps multi-view observations to the state representation expected by the expert, after which the original controller produces the action. This preserves the expert's control structure while restricting learning to the perception problem. Conceptually, the distinction is between learning an approximation of the complete mapping $o_t \mapsto a_t$ and learning only the missing state information $o_t \mapsto \hat{s}_t \mapsto \pi^{*}(\hat{s}_t)$.
The latter formulation can exploit the fact that the expert $\pi^{*}$ already provides a task-solving control policy and that the state representation imposes a structured intermediate target. Prior work has already combined state-conditioned experts with observation encoders~\cite{paolillo2022visualservoinggeometricallyinterpretable, andrychowicz2018learning, handa2023dextreme}. However, the expert policies are still susceptible to errors in the state prediction.

We propose a scheduled objective that interpolates from an estimation loss $L_s$, which bootstraps the state estimate, to a behaviour-cloning loss $L_a$ propagated through the expert (Fig.~\ref{fig:method}).
The gap that retaining the expert opens is large where the task is hard: on in-hand reorientation the estimator reaches 79\% success against under 1\% for the strongest pixel-to-action baseline, and a policy cloned from the true full state fails there as well, which points at the imitation objective rather than at perception.
The schedule in turn matters most where state supervision alone leaves the estimator furthest from the expert: on tray balancing it raises success from 29\% to 64\%.

Our contributions can be summarised as follows:
\begin{itemize}
\item We provide a study across diverse goal-conditioned manipulation tasks that
  compares direct action distillation with state-of-the-art imitation-learning
  methods against state estimation for a reused frozen expert, all trained on the
  same expert demonstration corpus, with each method's data filtering and
  regularisation set in its own favour and each imitation baseline quoted on its
  best of three vision backbones (Sec.~\ref{sec:baselines}).
\item We introduce a scheduled, mixed state-action loss to improve sufficient state estimation for the downstream expert.
  It combines direct state regression with an action-consistency loss backpropagated through the frozen, differentiable expert.
\item We ablate this objective against state-only, action-only, and fixed
  mixtures, and quantify the gap between state regression alone and the mixed
  objective.
\item We demonstrate sim-to-real transfer by deploying the frozen expert and a
  learned state estimator on physical hardware.

\end{itemize}

\section{Related Work}

Many manipulation policies deployed in the real world are trained on
demonstrations collected through teleoperation of real robots. ACT~\cite{zhaoact},
diffusion-based policies~\cite{chi2024diffusionpolicy} and large-scale generalist
vision--language--action models~\cite{pi05} learn observation--to--action mappings
from such data and generalise across tasks, but the cost of collecting it remains
one of the main bottlenecks of these methods~\cite{kawaharazuka2025vision}.

Simulation removes this data-collection cost. A policy trained on privileged simulator state
under domain randomisation~\cite{tobin2017domainrandomizationtransferringdeep} can be distilled
into a student that acts from onboard sensing~\cite{chen2020lbc,lee2020locomotion}, as in Visual
Dexterity, DextrAH-G and DextrAH-RGB~\cite{chen2023visualdexterity,lum2025dextrahg,singh2024dextrahrgb}.
In this line of research, the student replaces the teacher during deployment, so the control mapping is learned a second time.
State supervision already appears within this family. DextrAH-G adds an
object-position regression term to its distillation loss, though the predicted position
feeds a hand-designed state machine at deployment rather than a retained
controller~\cite{lum2025dextrahg}. Because the student visits states the teacher never demonstrated, these systems distil with
online DAgger~\cite{ross2011dagger}, querying the privileged teacher on states the student
itself reaches~\cite{chen2020lbc,lee2020locomotion,chen2023visualdexterity,lum2025dextrahg,singh2024dextrahrgb}.

A second line keeps the state-based controller and learns only a perceptual interface to it.
Position-based visual servoing is the early modular precedent, estimating object pose and passing
it to a control law defined on that pose~\cite{chaumette2006visualservoing}. OpenAI's dexterous
in-hand manipulation system trains a state-based policy, holds it fixed, and fits a vision network
by regression to the object pose that the policy consumes at deployment~\cite{andrychowicz2018learning},
and DeXtreme uses the same decomposition with a separately trained estimator substituted on the real
robot~\cite{handa2023dextreme}. Pitz et al.\ train an estimator for a previously learned policy but
then refine the controller with it in the loop~\cite{pitz2023dextrous}, and R\"ostel et al.\ report
that combining the two only at test time degrades performance~\cite{rostel2023estimatorcoupled}.
Closest to our setting, Huang et al.\ freeze a simulation-trained controller and train a bridge
module by matching expert actions through it, but that bridge targets the controller's learned
representation rather than physical state in known units~\cite{huang2026best}.

A third line of research also retains the controller but reconstructs learned privileged conditioning rather
than explicit physical state. Rapid Motor Adaptation trains a base policy together with a privileged
encoder and then learns an adaptation module predicting those latents from deployable observation
history~\cite{kumar2021rma}.
RotateIt reuses its base policy unchanged, replacing only the privileged encoder with a visuotactile transformer~\cite{qi2023rotateit}. There
the conditioning interface is learned jointly with the controller, so the reconstructed quantity is
a latent with shifting semantics. Our expert is pretrained independently, and its inputs
are semantically defined physical variables fixed before any perceptual module exists.

We study the approach of the mentioned second line of research in a
strict post-hoc setting, where the expert remains frozen throughout perceptual learning
and deployment. Prior systems in this family fit the estimator by supervised regression
against the ground-truth state. We ask whether that state interface should instead be
optimised for state fidelity, for action consistency through the frozen expert, or for
both, and compare this directly against distilling the same expert into a visuomotor
policy. Training a predictor against a fixed downstream objective rather than prediction error
alone is the subject of decision-focused learning~\cite{donti2017task}, where the
prediction and task terms are known to conflict and need balancing~\cite{bansal2023taskmet};
there the downstream module is a differentiable optimiser, here a frozen control policy
over physical variables.
\begin{figure*}[t]
    \centering
    \includegraphics[width=\linewidth]{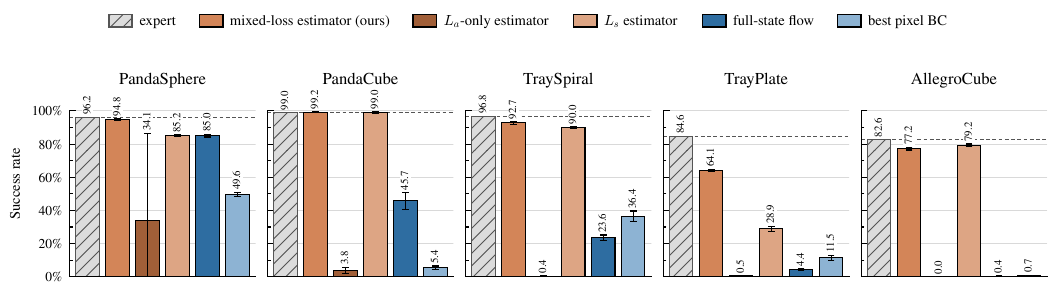}
    \caption{Closed-loop success rate per environment. Hatched: the frozen expert on the true state, repeated as a dashed reference. Orange: estimators with the retained expert (mixed-loss estimator with the linear ramp, $L_s$ estimator, $L_a$-only estimator). Blue: action imitation (full-state flow policy and the best pixel BC baseline per task over flow, ACT, MIP and three vision trunks). Bars are means over three training seeds, error bars the sample s.d. The pixel-BC bars are flow on ResNet-18, seeded on all five tasks; every other pixel-BC cell remains single-seed. The wide PandaSphere $L_a$ bar averages one converged and two collapsed seeds; protocol and values as in Tables~\ref{tab:full} and \ref{tab:mixed-loss-by-env}.}
    \label{fig:results-bars}
\end{figure*}

\section{Methodology}
\label{sec:method}

\subsection{Problem Setting and Privileged Expert}

We consider goal-conditioned manipulation tasks with a fixed goal $g$ per episode. A privileged expert $\teacher$ maps the state $\xs_t(g)$ to an action $\act_t=\teacher(\xs_t(g))$. The expert is goal-conditioned through the goal-relative entries of that state, which we make explicit by the argument $g$; it takes no separate goal input. We keep the argument implicit wherever no confusion arises. Training uses simulator access to state variables that are unavailable at deployment. The expert remains frozen and must be differentiable with respect to its input to train the estimator through the
  behaviour cloning loss. Details of the experts used in our experiments are given in
  Sec.~\ref{sec:experiments}.

The expert input $\xs_t(g)$ has two parts. The measured part $s_{q,t}$, such as joint positions and velocities, is measured by the robot's own sensors and is therefore also available at deployment. The remaining part $s_{o,t}(g)$, the non-proprioceptive state, comprises the object state, such as its position, tracked keypoints and velocity, and the goal-relative quantities derived from it, such as the goal position minus the object position, so that $\xs_t(g)=(s_{q,t},\,s_{o,t}(g))$. It is read from the simulator during training but is not measured at deployment, where it must be inferred from camera images.

We use the expert to collect a demonstration set $\mathcal{D}=\{(\obs_t,\xs_t,\act_t,g)\}$ by rolling it out in simulation on the true state, so that every sample provides a state target $s_{o,t}$ and an action target $\act_t=\teacher(\xs_t(g))$ for the same input. Rather than cloning the expert's actions from pixels, we retain $\teacher$ as the controller and learn only an estimator $\phis$ for its unobserved input $s_{o,t}$.

\subsection{Visual State Estimation and Retained Control}
The estimator $\phis$ maps the last $H$ camera images $\obs_{t-H+1:t}$, the measurements $s_{q,t-H+1:t}$ over the same window, and the goal $g$ to an estimate $\hat{s}_{o,t}$ of the current non-proprioceptive state. A shared visual backbone, a LoRA-adapted DINOv3 here (Sec.~\ref{sec:baselines}), encodes each image. The image features of every frame are concatenated with that frame's proprioception; a temporal convolutional network aggregates the $H$ frames; and an MLP that additionally receives $g$ outputs $\hat{s}_{o,t}$. Since $g$ is an input, the goal-relative entries of $\hat{s}_{o,t}$ are predicted directly rather than computed as the goal minus a predicted object position: they are keypoint coordinates relative to a task reference point, so the two blocks are free to disagree and we do not enforce their consistency. Whether constructing them analytically helps is untested.

The expert input is then the concatenation of the measured proprioception and the estimate, in the expert's input order,
\begin{equation}
  \est_t = \left(s_{q,t},\,\hat{s}_{o,t}\right),
\end{equation}
so that $\est_t$ differs from $\xs_t(g)$ only in the estimated entries. The deployed action is
\begin{equation}
  \actp_t = \teacher(\est_t(g)).
\end{equation}
The values for $H$, the visual backbone and all optimisation hyperparameters are given in Sec.~\ref{sec:experiments}.

\subsection{Training Objectives}
We train the estimator with two losses, one defined on states and one on actions. The \emph{estimation loss} $\Ls$ compares the predicted state $\hat{s}_{o,t}$ with the true state $s_{o,t}$; the \emph{behaviour cloning loss} $\La$, abbreviated BC loss below, compares the action the frozen expert produces from that prediction with the recorded expert action (Fig.~\ref{fig:method}). The estimation loss runs over the $d$ entries of the non-proprioceptive state, each divided by the standard deviation $\sigma_i$ of $[s_{o,t}]_i$ in the demonstration set,
\begin{equation}
  \Ls = \mathbb{E}_t\!\left[\frac{1}{d}\sum_{i=1}^{d}
    \left(\frac{[\hat{s}_{o,t}-s_{o,t}]_i}{\sigma_i}\right)^2\right].
\end{equation}
This is the MSE between the prediction and the target in these normalised units, over the whole non-proprioceptive state including its goal-relative entries. Proprioception and the goal are inputs, not regression targets.

The BC loss compares the expert's response to this estimated input with its recorded demonstration action,
\begin{equation}
  \La = \mathbb{E}_t\!\left[\|\teacher(\est_t(g))-\act_t\|_2^2\right].
\end{equation}
The expert's parameters remain fixed, but backpropagating through the expert lets this loss update the estimator.

We combine the two objectives directly,
\begin{equation}
  \mathcal{L}(p)=\ws(p)\Ls+\wa(p)\La,
\end{equation}
where $p\in[0,1]$ denotes training progress. The state targets remain normalised as defined above; no additional loss normalisation is applied.

\paragraph{The BC Loss as Re-Weighted State Estimation}
The expert is fixed but differentiable with respect to its state input. Since the proprioceptive entries of $\est_t$ are supplied, only its Jacobian with respect to the estimated entries,
\begin{equation}
  J_{\pi,t}=\frac{\partial\teacher(\est_t)}{\partial\hat{s}_{o,t}},
\end{equation}
enters the gradient of $\La$, no expert parameter is optimised and no derivative through the simulator is required.

The expert is deterministic, so $\act_t=\teacher(\xs_t(g))$, and proprioception is supplied exactly, so $\est_t$ and $\xs_t(g)$ differ only by the estimation error $e_{s,t}=\hat{s}_{o,t}-s_{o,t}$. To first order in $e_{s,t}$, the action error is $\teacher(\est_t)-\teacher(\xs_t(g))\approx J_{\pi,t}\,e_{s,t}$, and therefore
\begin{equation}
\label{eq:action-metric}
  \La\approx\mathbb{E}_t\!\left[e_{s,t}^{\top}M_t\,e_{s,t}\right],\qquad
  M_t=J_{\pi,t}^{\top}J_{\pi,t}.
\end{equation}
Training on $\La$ is thus a re-weighted version of training on $\Ls$. Locally the two are the same quadratic in the estimation error, with the diagonal metric of $\Ls$ replaced by the state-dependent metric $M_t$ induced by the expert. $M_t$ describes which aspects of the state are relevant to estimate for effective imitation. Directions to which the expert is locally insensitive receive little or no first-order penalty, whereas directions that change its action dominate the loss. In this sense the objective realises a form of \emph{sufficient}, or relevance-reweighted, state estimation: the estimate must be accurate where accuracy affects control, not uniformly across coordinates.

\paragraph{Schedules}
The metric $M_t$ is only as good as the linearisation behind it: $J_{\pi,t}$ is evaluated at the estimate, so $M_t$ is informative once $\est_t$ lies near states on which the expert was trained. We therefore begin with direct state supervision and increase the contribution of the BC loss over training. Our schedules are convex combinations, $\ws(p)=1-\wa(p)$, so each is defined by $\wa$ alone, rising from $0$, pure estimation supervision, to $1$, pure BC supervision. Our default is the linear ramp
\begin{equation}
  \wa(p)=p,
\end{equation}
which moves continuously from estimation to BC supervision. We also evaluate the cosine ramp
\begin{equation}
  \wa(p)=\frac{1-\cos(\pi p)}{2},
\end{equation}
and a staged switch, $\wa(p)=0$ for $p<p_0$ and $1$ thereafter, with $p_0=0.5$. As fixed-weight baselines, which are not convex combinations, we use $\ws=1$ and $\wa=\lambda$ and sweep $\lambda$ to vary the relative contribution of BC supervision, and as a limiting case we train on the action loss alone, $\ws=0$ and $\wa=1$ throughout.

\section{Experimental Setup}
\label{sec:experiments}

\subsection{Tasks and Demonstration Collection}
We evaluate on five goal-conditioned manipulation tasks (Fig.~\ref{fig:environments}), three of them adapted from the Constrained Sampling and Reinforcement Learning (CSRL) benchmark~\cite{toussaint2026combinedconstrainedsamplingreinforcement}. In PandaSphere, a Panda arm brings a free sphere to a goal position; the goals are stable configurations in which the sphere rests against one of the robot's links, so the task calls for whole-body contact rather than grasping. In PandaCube, the arm pushes a cube lying flat on a table to a goal position and yaw, in a slower variant of the original task. In AllegroCube, a 16-DoF Allegro hand reorients a cube in-hand to a goal pose. In the two tray tasks the arm tilts and moves a held tray until each of two spheres reaches its own goal position without falling off; TraySpiral adds a spiral wall maze, TrayPlate is an open tray. Success is defined identically across tasks: each tracks a set of keypoints---the sphere centre in PandaSphere, the cube centre and its four corners in PandaCube and AllegroCube, the two sphere centres in the tray frame on the trays---and an episode succeeds, and ends, at the first step at which the Euclidean norm of the stacked keypoint error falls below a tolerance $\epsilon$, with $\epsilon=1$\,cm for PandaSphere, $5$\,cm for PandaCube and $2$\,cm for the remaining three tasks. The corner keypoints couple position and orientation, so no separate angular tolerance is needed, and the joint norm makes $\epsilon$ a budget shared across keypoints rather than a per-point threshold. Table~\ref{tab:state-composition} gives the composition of the expert input: the estimator predicts object positions and velocities, for the cubes four corner markers in place of a quaternion, and the same keypoints relative to their goal; the remainder, including the AllegroCube fingertips, is measured. Actions are increments of the joint position reference.

\begin{table}[t]
  \centering
  \caption{Expert input per task. The estimator predicts only $s_o$, the object and goal-relative
  entries; $s_q$ is measured at deployment.}
  \label{tab:state-composition}
  \small
  \setlength{\tabcolsep}{3.5pt}
  \begin{tabular}{@{}lccccc@{}}
    \toprule
    & & measured & \multicolumn{2}{c}{estimated $s_o$} & \\
    \cmidrule(lr){4-5}
    Task & $\dim s$ & $s_q$ & object & goal-rel. & $\dim a$ \\
    \midrule
    PandaSphere          & 42 & 30 &  9 &  3 &  7 \\
    PandaCube            & 66 & 30 & 21 & 15 &  7 \\
    TraySpiral, TrayPlate & 57 & 33 & 18 &  6 &  7 \\
    AllegroCube          & 96 & 60 & 21 & 15 & 16 \\
    \bottomrule
  \end{tabular}
\end{table}

We train the privileged experts with CSRL~\cite{toussaint2026combinedconstrainedsamplingreinforcement}, using TD7~\cite{fujimoto2023td7} with a HER term~\cite{andrychowicz2017her} in MuJoCo Warp~\cite{mujocowarp2025}: 10,000 stable states, configurations in equilibrium under external forces, serve as reset states and goals, and a curriculum progressively increases the time horizon between them. Experts train for 200,000 steps.

We then roll out each expert for 10,000 demonstration episodes from the same stable-state set, recording images, simulator state and expert action at every step. PandaSphere and AllegroCube use two external cameras, PandaCube three, and the tray tasks one external and one wrist-mounted camera. All methods train on this corpus; the per-method filtering of it is stated in Sec.~\ref{sec:baselines}.

\begin{table*}[t]
  \centering
  \caption{Closed-loop success (\%), protocol as in Sec.~\ref{sec:experiments}. $\pm$ is the sample s.d.\ over three training seeds; cells without it are single-seed and are the best of three trunks, the superscript naming the winner (\textsuperscript{R18}~ResNet-18, \textsuperscript{v2}~DINOv2-LoRA, \textsuperscript{v3}~DINOv3-LoRA)\ifwithappendix; every trunk is listed in Table~\ref{tab:imitation-trunks}\fi. The flow column is ResNet-18 throughout, so it carries no superscript. Both estimators use DINOv3-LoRA, the mixed-loss estimator the linear ramp. \textbf{Bold} = best cell per row excluding the expert.}
  \label{tab:full}
  \small
  \begin{tabular}{@{}lccc c cc c@{}}
    \toprule
    & \multicolumn{3}{c}{\emph{pixels $\rightarrow$ actions}} & \emph{state $\rightarrow$ actions} & \multicolumn{2}{c}{\emph{pixels $\rightarrow$ state $\rightarrow$ actions}} & \\
    \cmidrule(lr){2-4}\cmidrule(lr){5-5}\cmidrule(lr){6-7}
    Task & flow & ACT & MIP & full-state flow & $L_s$ estimator & mixed-loss estimator & expert \\
    \midrule
    \textsc{PandaSphere} & 49.6$\pm$1.3 & 43.5\textsuperscript{R18} & 29.2\textsuperscript{v3} & 85.0$\pm$0.8 & 85.2$\pm$0.3 & \textbf{94.8$\pm$0.4} & 96.2 \\
    \textsc{PandaCube} & 5.4$\pm$1.1 & 6.1\textsuperscript{R18} & 3.5\textsuperscript{R18} & 45.7$\pm$5.1 & 99.0$\pm$0.4 & \textbf{99.2$\pm$0.4} & 99.0 \\
    \textsc{TraySpiral} & 36.4$\pm$3.2 & 21.8\textsuperscript{R18} & 32.0\textsuperscript{v3} & 23.6$\pm$1.6 & 90.0$\pm$0.5 & \textbf{92.7$\pm$0.8} & 96.8 \\
    \textsc{TrayPlate} & 11.5$\pm$1.5 & 10.7\textsuperscript{R18} & 2.5\textsuperscript{v3} & 4.4$\pm$0.5 & 28.9$\pm$1.7 & \textbf{64.1$\pm$0.7} & 84.6 \\
    \textsc{AllegroCube} & 0.7$\pm$0.2 & 0.6\textsuperscript{v3} & 0.5\textsuperscript{v3} & 0.4$\pm$0.1 & \textbf{79.2$\pm$0.8} & 77.2$\pm$0.9 & 82.6 \\
    \bottomrule
  \end{tabular}
\end{table*}

\subsection{Baselines and Implementation}
\label{sec:baselines}
Throughout, the \emph{$L_s$ estimator} is the retained-expert estimator trained
with the estimation loss alone, and the \emph{mixed-loss estimator} the same
estimator trained with $w_s(p)L_s+w_a(p)L_a$, combined without normalisation. We compare both with three pixel-to-action imitation
baselines. Action Chunking with Transformers (ACT)~\cite{zhaoact} drops its
variational encoder, since the
demonstrations are deterministic. The flow-matching
policy~\cite{lipman2023flowmatching} follows the action-chunking formulation of
Diffusion Policy~\cite{chi2024diffusionpolicy} with a diffusion-transformer
denoiser~\cite{dasari2025ingredients}. The Minimum Iterative Policy
(MIP)~\cite{pan2026adonoisingdispellingmyths} replaces iterative sampling by a direct regression of the action chunk and one
refinement pass. The \emph{full-state flow} policy clones the expert with the
same flow-matching head from the true full state $\xs_t$ and the goal, without
images, and therefore has strictly more information than any pixel-based method.

\paragraph{Shared Pipeline}
Every pixel-based method receives the same camera frames $\obs_{t-H+1:t}$,
measurements $s_{q,t-H+1:t}$ and goal $g$, with $H=4$. Images are encoded by one
of three trunks, a ResNet-18~\cite{he2015deepresiduallearningimage} trained from
scratch or a DINOv2~\cite{oquab2024dinov2} or DINOv3~\cite{simeoni2025dinov3}
ViT-B/16 with rank-16 LoRA~\cite{hu2022lora} on the query, key and value
projections of the last four blocks and all other backbone parameters fixed. The
trunk is shared across cameras and timesteps, and spatial-softmax pooling
followed by a two-layer adapter yields a 256-dimensional representation per
image. Every imitation objective is trained on all three trunks\ifwithappendix\ (Table~\ref{tab:imitation-trunks})\fi. ResNet-18 is the
best trunk in ten of the fifteen imitation cells of Table~\ref{tab:full}; on
\textsc{AllegroCube} the estimator spans 1.6 points across the three trunks, and it
uses DINOv3 throughout.

All methods train for 40,000 steps at an initial learning rate of
$2\times10^{-4}$ with cosine decay, in mixed precision.

\paragraph{Heads}
The methods differ only after the trunk. In the estimator, the camera
representations of each timestep are concatenated with the proprioception, a
temporal convolution aggregates the four-frame window, and an MLP conditioned on
the goal predicts the standardised object-state and goal-relative entries of the
last frame, which are decoded and composed with the true proprioception as in
Sec.~\ref{sec:method}. In the baselines, embedded proprioception and goal
condition an eight-block, width-256 diffusion transformer through adaptive layer
normalisation, which the flow head integrates in 10 Euler steps and MIP queries
at refinement time $0.9$; ACT keeps the LeRobot defaults with chunk size two and
no temporal ensembling. Actions are min--max normalised, and every policy
predicts two consecutive actions and executes the first---the best of six horizon
settings in a sweep on PandaSphere.

\paragraph{Data and Regularisation}
Two settings differ between the pipelines, each chosen in its own favour. The
baselines train on successful demonstrations only, since cloning failed actions
degrades imitation, and apply proprioception dropout at 0.15 per sample. The
estimator trains on the full corpus, since a failed episode still carries
correct states, and uses neither.

\subsection{Evaluation Protocol}
We report closed-loop success rates over 1,000 episodes with a fixed evaluation seed, in the rendering in which the demonstrations were collected, with domain randomisation disabled, at the final checkpoint of the fixed training budget. The $L_s$-only, linear-ramp, cosine-ramp and $L_a$-only estimators and the full-state flow policy are trained with three seeds on every task and reported as mean and sample standard deviation across seeds. Among the pixel imitation baselines, flow on ResNet-18, the strongest single-seed configuration, is additionally given three seeds; the staged and flat mixed-loss experiments and every other pixel-BC cell stay single-seed, so we do not claim significance against them. Expert reference rates are measured with the same harness.

\subsection{Real-World Experimental Setup}
\label{sec:real-setup}
\begin{figure*}[t]
  \centering
  \setlength{\drlab}{9pt}\setlength{\drgap}{2pt}\setlength{\drsep}{8pt}
  \def\drscale{1}%
  \setlength{\drtile}{\drscale\dimexpr(\textwidth-\drlab-5\drgap-2\drsep-1pt)*2/17\relax}
  \drhead{\drlab}{}\drhead{\dimexpr3\drtile+2\drgap\relax}{simulation, domain-randomised}\hspace{\drsep}\drhead{\drtile}{real}\hspace{\drsep}\drhead{\dimexpr4\drtile+3\drgap+\drtile/2\relax}{real-world setup}\\[1pt]
  \begin{minipage}[b]{\dimexpr\drlab+4\drtile+2\drgap+\drsep\relax}\centering
    \drlabel{cam 0}\includegraphics[width=\drtile]{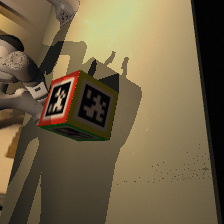}\hspace{\drgap}\includegraphics[width=\drtile]{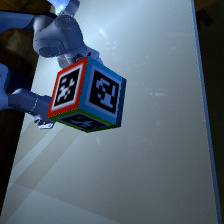}\hspace{\drgap}\includegraphics[width=\drtile]{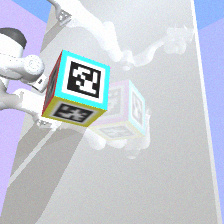}\hspace{\drsep}\includegraphics[width=\drtile]{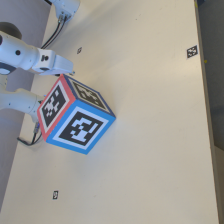}\\[\drgap]
    \drlabel{cam 1}\includegraphics[width=\drtile]{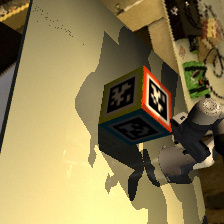}\hspace{\drgap}\includegraphics[width=\drtile]{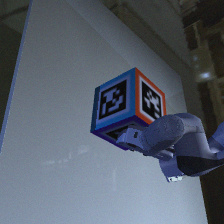}\hspace{\drgap}\includegraphics[width=\drtile]{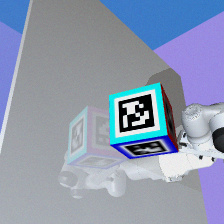}\hspace{\drsep}\includegraphics[width=\drtile]{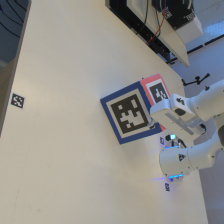}\\[\drgap]
    \drlabel{cam 2}\includegraphics[width=\drtile]{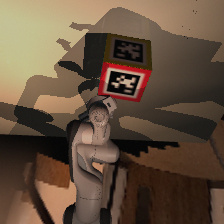}\hspace{\drgap}\includegraphics[width=\drtile]{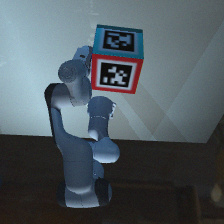}\hspace{\drgap}\includegraphics[width=\drtile]{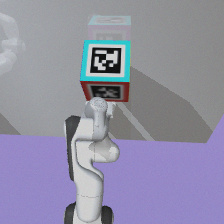}\hspace{\drsep}\includegraphics[width=\drtile]{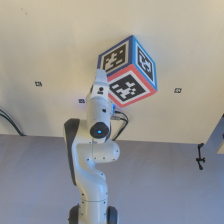}
  \end{minipage}\hspace{\drsep}%
  \begin{minipage}[b]{\dimexpr4\drtile+3\drgap+\drtile/2\relax}\centering
    \includegraphics[height=\dimexpr3\drtile+2\drgap\relax]{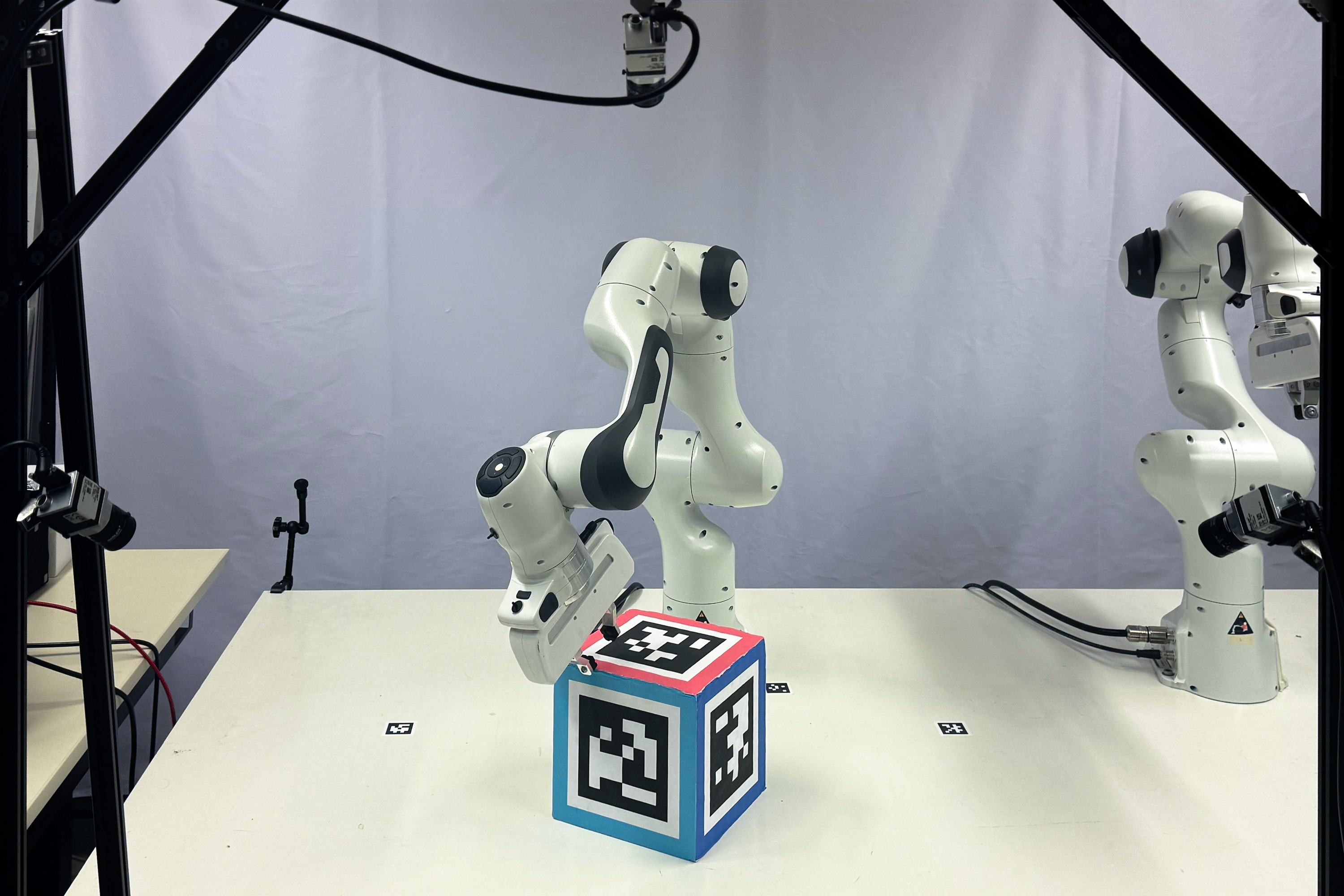}
  \end{minipage}
  \caption{Domain-randomised training observations and real-world setup for PandaCube. Rows are the three cameras. Left: three simulated training episodes; middle: real-world observations; right: the physical setup.}
  \label{fig:dr-episodes}
\end{figure*}

We evaluate sim-to-real transfer on a physical Panda setup (Fig.~\ref{fig:dr-episodes}, right) on the PandaCube task. With calibrated cameras, we collect 10,000 simulated demonstration episodes with visual domain randomisation~\cite{tobin2017domainrandomizationtransferringdeep} of the background (procedural rooms, textures, checkerboards, noise and gradients), the positions, intensities and colour temperatures of two lights, material reflectance and cube frame colours (Fig.~\ref{fig:dr-episodes}). The deployed policy is the mixed-loss estimator with the linear ramp; this domain-randomised training line is reported separately from the simulation experiments above.

We run 25 episodes, each starting from a nominal robot configuration and a fixed cube start position whose yaw advances by $45^\circ$ per episode. The goal is drawn at random from the stable-configuration set and the policy has 40\,s to reach it. AprilTags on the cube serve solely to detect episode termination; the tracked pose is not supplied to the estimator or the expert.

\section{Results}
\label{sec:results}

\subsection{State-Estimation versus Action Imitation}
Figure~\ref{fig:results-bars} and Table~\ref{tab:full} compare the $L_s$ and mixed-loss estimators with the imitation baselines and the full-state flow policy.
At the shared training budget, the mixed-loss estimator outperforms every baseline on every task, including the full-state flow policy, and the $L_s$ estimator does so everywhere except PandaSphere, where it matches full-state flow (85.2 vs.\ 85.0\%); the margins reach over seventy points on AllegroCube, where both estimators approach the expert while even cloning from the true full state fails. The margin is task-dependent rather than a universal failure of action cloning.

\subsection{Loss Weighting and Scheduling}
Table~\ref{tab:mixed-loss-by-env} compares the $L_s$-only base with staged switches, linear and cosine ramps, flat weights $w_a\in\{1,0.1\}$, and pure action-loss training.

On PandaSphere, the staged switch, both ramps and the flat unit-weight experiment reach similar success rates, all about nine points above the $L_s$-only baseline. This agreement does not extend to the tray tasks. There, the cosine ramp has the highest success, the staged switch and the flat unit-weight experiment collapse to below 4\%, far under the $L_s$-only baseline, and reducing the flat action weight to 0.1 recovers performance close to the ramps. On PandaCube the baseline already matches the expert and every schedule stays within a point of it. On AllegroCube no schedule improves on the $L_s$-only baseline. Both ramps fall one to two points below it, and the flat unit-weight experiment collapses. Schedule and relative loss weighting thus materially affect the estimator. The two ramp shapes, by contrast, differ by at most 2.2 points on any task, comparable to the spread across training seeds; we therefore report the linear ramp throughout, in simulation and on the real robot. At a single seed, flat $w_a{=}0.1$ also stays within 1.9 points of the linear ramp on every task; the arms that fail are those trained at $w_a{=}1$ throughout or switched to it abruptly. We report the ramps because they reach full action supervision without a per-task weight to tune.

\begin{table*}[t]
  \centering
  \caption{Mixed-loss schedules. Success rate (\%), protocol as in Table~\ref{tab:full}.
           \emph{expert}: the frozen expert on the true state; \emph{$L_s$ only}: the
           state-only estimator, the baseline. \textbf{Bold} = best schedule per row, \underline{underline} = below that baseline. The PandaSphere $L_a$-only entry
           averages one converged seed (94.6) with two collapsed ones (1.9, 5.9);
           see Sec.~\ref{sec:results-interpretation}.}
  \label{tab:mixed-loss-by-env}
  \small
  \setlength{\tabcolsep}{6pt}
  \begin{tabular}{@{}lc c cc cc cc@{}}
    \toprule
      & & staged & \multicolumn{2}{c}{ramped} & \multicolumn{2}{c}{flat} & & \\
    \cmidrule(lr){3-3}\cmidrule(lr){4-5}\cmidrule(lr){6-7}
    Task & expert & switch & linear & cosine
      & $w_a{=}1$ & $w_a{=}0.1$ & $L_s$ only & $L_a$ only \\
    \midrule
    \textsc{PandaSphere} & 96.2 & 94.6 & \textbf{94.8$\pm$0.4} & 94.5$\pm$0.3 & 93.6 & 92.9 & 85.2$\pm$0.3 & \underline{34.1$\pm$52.4} \\
    \textsc{PandaCube} & 99.0 & \textbf{99.5} & 99.2$\pm$0.4 & 99.2$\pm$0.3 & \underline{98.4} & 99.4 & 99.0$\pm$0.4 & \underline{3.8$\pm$2.0} \\
    \textsc{TraySpiral} & 96.8 & \underline{1.0} & 92.7$\pm$0.8 & \textbf{94.3$\pm$1.2} & \underline{3.8} & 93.9 & 90.0$\pm$0.5 & \underline{0.4$\pm$0.2} \\
    \textsc{TrayPlate} & 84.6 & \underline{0.8} & 64.1$\pm$0.7 & \textbf{66.3$\pm$1.8} & \underline{0.8} & 62.8 & 28.9$\pm$1.7 & \underline{0.5$\pm$0.2} \\
    \textsc{AllegroCube} & 82.6 & \textbf{\underline{79.1}} & \underline{77.2$\pm$0.9} & \underline{78.1$\pm$0.8} & \underline{8.7} & \underline{78.3} & 79.2$\pm$0.8 & \underline{0.0$\pm$0.0} \\

    \bottomrule
  \end{tabular}

\end{table*}

\subsection{Interpretation of the Results}
\label{sec:results-interpretation}
The reported success rates do not isolate the causes behind these differences; we discuss plausible explanations.

\paragraph{Why Action Imitation Falls Behind}
A goal-conditioned clone has to reproduce a family of feedback behaviours, and a visuomotor clone additionally has to learn visual state inference, from rollouts of a deterministic expert that cover recovery behaviours sparsely~\cite{laskey2017dart}. With a fixed demonstration set, errors in the cloned action move the clone into poorly covered states and compound over time~\cite{ross2011dagger}. Retaining the expert preserves its feedback law and limits learning to estimating its inputs. At the shared budget, cloning from the true full state falls short on all five tasks, most starkly on AllegroCube (0.4\% vs.\ 82.6\%), so this matters without perception. With three times the budget it closes much of the gap on the reaching and tray tasks but not on AllegroCube\ifwithappendix\ (Appendix~\ref{app:mlp})\fi. The pipelines differ in data filtering and regularisation (Sec.~\ref{sec:baselines}), each in its own favour.
A plain MLP cloned from the true state confirms this: on AllegroCube it stays below 7\% at every capacity up to four times the expert's size and at three times the training budget, although the demonstrated action is an exact function of that state\ifwithappendix\ (Appendix~\ref{app:mlp})\fi.

\paragraph{Spatial Representation on the Tray Tasks}
On the tray tasks, pixel imitation outperforms the full-state flow policy (Table~\ref{tab:full}). The state vector holds the spheres' coordinates and goals but not the geometry the spheres have to be steered through, which a policy cloned from it has to infer from the demonstrations alone, whereas the images expose it at every step. The original expert acquired this structure during reinforcement learning and keeps it, so an estimator paired with it only has to recover the sphere coordinates. This hypothesis is not isolated from differences in architecture, optimisation or temporal inputs. With three times the training budget the full-state flow policy reaches 83.2\% on TraySpiral and 36.5\% on TrayPlate, so a large part of this gap reflects slower convergence rather than missing information.

\paragraph{Why Combine State and Action Supervision}
State regression penalises coordinate errors equally in standardised units, whereas action supervision emphasises the errors that change the expert's output (Eq.~\ref{eq:action-metric}); the two are complementary, one grounding the prediction in the physical state, the other making it sensitive to downstream control. The benefit is largest where state-only training is far from the expert, on TrayPlate (28.9\% to 64.1\% with the linear ramp), and small where it is already close, on TraySpiral (90.0\% to 92.7\%); Fig.~\ref{fig:results-bars} shows the pattern across all five tasks.

\paragraph{Why Action Loss Alone Can Fail}
Matching the expert's action need not determine the state uniquely, but the PandaSphere runs point to an optimisation failure instead. Of three $L_a$-only seeds, one converges to 94.6\% success, level with the ramps, while the other two stay below 6\%, the source of the large deviation in Table~\ref{tab:mixed-loss-by-env}; no other action-only run on any task converges. On held-out demonstrations, the collapsed seeds mispredict the goal-relative object position by metres, whereas the converged seed predicts it more accurately than the $L_s$ estimator (median error 0.83 versus 1.37\,cm) and nearly as accurately as a ramp-trained estimator (0.70\,cm). The BC loss thus defines a good objective whose minimum is hard to reach from a random initialisation; state supervision conditions the optimisation, placing the estimates where gradients through the expert become useful, after which a ramp emphasises control-relevant errors. Because the ramps end with action supervision alone, this does not establish that the estimation loss must remain active throughout training.

\subsection{Real-World Transfer}
\label{sec:real-results}
Using the protocol in Sec.~\ref{sec:real-setup}, the mixed-loss estimator trained with the linear ramp reaches 19 of 25 goals within the time budget (76\%). We classify the six failures into two modes: four episodes run out of time, and in two the cube is pushed out of the workspace. Across all episodes the dominant difference to simulation is a dynamics mismatch, with physical parameters that differ from the simulated ones, the real cube responds differently to the same actions and is turned too far or not far enough. This shows most in the policy's strategy for sharp rotations, pivoting the cube on a corner, often with the edge of the wrist housing, where small errors in contact geometry and estimated state produce repeated motions without progress. The policy usually recovers, by re-establishing contact or trying a slightly different motion, at the cost of time. Real episodes take $78.2 \pm 45.8$ control steps against $26.1 \pm 12.5$ in simulation, where the same policy succeeds in 99.2\% of episodes (Table~\ref{tab:sim-real}). The polystyrene cube also degrades over the experiments, with dented, rounded corners, so the object is not identical across episodes---a substantial deviation for a corner-pivoting strategy.
\begin{table}[t]
  \centering
  \caption{Sim-to-real evaluation of the domain-randomised policy.}
  \label{tab:sim-real}
  \small
  \setlength{\tabcolsep}{3pt}
  \begin{tabular}{@{}llcc@{}}
    \toprule
    Domain & Objective & Success (\%) & Mean steps \\
    \midrule
    Simulation & Mixed loss, linear ramp & 99.2 & $26.1 \pm 12.5$ \\
    Real world & Mixed loss, linear ramp & 76.0 & $78.2 \pm 45.8$ \\
    \bottomrule
  \end{tabular}
  \par\smallskip
  \begin{minipage}{\columnwidth}
    \footnotesize
    Real-world evaluation: 25 episodes, 40\,s per episode (a 200-step budget),
    out of 27 attempts;
    two were stopped by the controller's joint-limit safety check and restarted,
    and counting those as failures gives 70.4\%. Simulation: 1,000 episodes in a held-out
    draw of the domain randomisation, with background panoramas unseen in training. Mean steps: episode length in control
    steps, mean $\pm$ s.d.\ over successful episodes.
  \end{minipage}
\end{table}

\section{Conclusion}

This work studies the deployment of privileged state-based manipulation policies from visual observations without replacing the original expert. Instead of end-to-end distillation, we retain the frozen expert and learn only a perceptual module that estimates its unobserved state inputs. Across five goal-conditioned manipulation tasks, this decomposition consistently outperforms direct pixel-to-action imitation from the same demonstrations. We further show that combining direct state regression with action supervision through the differentiable expert can improve performance by emphasising state errors that matter for downstream control, although the optimal weighting remains task-dependent.

Finally, we demonstrate sim-to-real transfer on PandaCube, achieving 76\% success on the physical robot without retraining the underlying expert. The remaining failures are largely associated with dynamics and contact mismatch, motivating future work on physical domain randomisation, uncertainty-aware state estimation, and adaptation to model mismatch. Two further directions follow: a dedicated study isolating where and why pixel imitation falls behind, and tasks with faster, less recoverable dynamics.

\ifwithappendix\section*{Appendix}

\begin{table}[t]
  \centering
  \caption{Pixel imitation across vision trunks and objectives. Closed-loop success (\%), protocol as in Sec.~\ref{sec:experiments}, single training seed. Bold marks the best trunk within each task and objective; Table~\ref{tab:full} quotes the bold ACT and MIP cells, and for flow the ResNet-18 trunk averaged over three training seeds.}
  \label{tab:imitation-trunks}
  \small
  \begin{tabular}{@{}llccc@{}}
    \toprule
    Task & Trunk & flow & ACT & MIP \\
    \midrule
    \textsc{PandaSphere} & ResNet18 & \textbf{51.1} & \textbf{43.5} & 14.9 \\
     & DINOv2-LoRA & 20.7 & 26.0 & 10.8 \\
     & DINOv3-LoRA & 38.1 & 42.5 & \textbf{29.2} \\
    \midrule
    \textsc{PandaCube} & ResNet18 & \textbf{5.3} & \textbf{6.1} & \textbf{3.5} \\
     & DINOv2-LoRA & 2.4 & 1.6 & 0.7 \\
     & DINOv3-LoRA & 3.7 & 4.2 & 3.2 \\
    \midrule
    \textsc{TraySpiral} & ResNet18 & \textbf{37.8} & \textbf{21.8} & 15.3 \\
     & DINOv2-LoRA & 18.3 & 16.5 & 10.4 \\
     & DINOv3-LoRA & 28.8 & 18.5 & \textbf{32.0} \\
    \midrule
    \textsc{TrayPlate} & ResNet18 & \textbf{11.2} & \textbf{10.7} & 1.7 \\
     & DINOv2-LoRA & 2.2 & 3.1 & 1.6 \\
     & DINOv3-LoRA & 8.6 & 8.2 & \textbf{2.5} \\
    \midrule
    \textsc{AllegroCube} & ResNet18 & \textbf{0.9} & 0.2 & 0.2 \\
     & DINOv2-LoRA & 0.5 & 0.5 & 0.2 \\
     & DINOv3-LoRA & 0.6 & \textbf{0.6} & \textbf{0.5} \\
    \bottomrule
  \end{tabular}
\end{table}

\subsection{Is the Imitation Gap an Implementation Artefact?}
\label{app:mlp}
As a control for the low cloning cells of Table~\ref{tab:full}, we clone the expert from the true
state $\xs_t$ with a plain ReLU MLP trained by mean-squared error, removing the flow head, sampler,
transformer and action chunk; data, optimiser and schedule are those of the imitation baselines
(Sec.~\ref{sec:baselines}). Four widths bracket the expert's own size ($823$--$867$k parameters
where measured). On PandaSphere and AllegroCube, running the frozen expert on the recorded states
reproduces the demonstrated actions to float32 round-off, so an exact state-to-action clone exists.
Table~\ref{tab:mlp-control} reports the MLP and the full-state flow policy at the shared budget and
at three times it. Neither head is uniformly stronger: flow leads on PandaSphere and the trays, the
MLP on PandaCube. Both improve substantially with budget on every task except AllegroCube, where
nothing exceeds $7.0\%$ against an expert at $82.6\%$. At 120k steps full-state flow exceeds the
$\Ls$ estimator of Table~\ref{tab:full} on PandaSphere ($93.3$ vs.\ $85.2\%$) and TrayPlate
($36.5$ vs.\ $28.9\%$), while the mixed-loss estimator, trained at the shared budget, stays ahead
on all five tasks, by $1.5$ points on PandaSphere. The comparisons of Sec.~\ref{sec:results} are
therefore at matched budget, not at convergence.

\begin{table}[t]
  \centering
  \caption{Closed-loop success (\%) of a privileged-state MLP clone and the full-state flow policy
  at the shared 40k-step budget and at 120k steps; protocol as in Sec.~\ref{sec:experiments}.
  MLP widths relative to the frozen expert. $\pm$ = s.d.\ over three training seeds, other cells
  single-seed. \textsuperscript{$\dagger$}four-frame window and goal input, as for the imitation
  baselines; the other arms take one state frame. Bold = best per row.}
  \label{tab:mlp-control}
  \footnotesize
  \setlength{\tabcolsep}{2.4pt}
  \begin{tabular}{@{}llccccc@{}}
    \toprule
    & & \multicolumn{4}{c}{MLP} & full-state \\
    \cmidrule(lr){3-6}
    Task & steps & $0.18\times$ & $0.99\times$ & $0.2$--$0.3\times$\textsuperscript{$\dagger$} & $4.1\times$\textsuperscript{$\dagger$} & flow \\
    \midrule
    \textsc{PandaSphere} & 40k  & 13.8 & 63.7$\pm$1.9 & 23.7 & 63.4 & \textbf{85.0$\pm$0.8} \\
                         & 120k & 56.6 & 82.3 & 57.6 & 66.2 & \textbf{93.3} \\
    \midrule
    \textsc{PandaCube}   & 40k  & 9.3 & 59.3 & 8.3 & \textbf{80.7} & 45.7$\pm$5.1 \\
                         & 120k & 49.8 & \textbf{88.1} & 45.7 & 87.8 & 86.8 \\
    \midrule
    \textsc{TraySpiral}  & 40k  & 8.0 & 16.8 & 10.6 & 21.6 & \textbf{23.6$\pm$1.6} \\
                         & 120k & 22.5 & 28.1 & 18.4 & 26.0 & \textbf{83.2} \\
    \midrule
    \textsc{TrayPlate}   & 40k  & 0.7 & 1.2 & 1.1 & 4.0 & \textbf{4.4$\pm$0.5} \\
                         & 120k & 2.5 & 4.8 & 3.3 & 6.5 & \textbf{36.5} \\
    \midrule
    \textsc{AllegroCube} & 40k  & 0.5 & 0.5 & 0.1 & \textbf{1.5$\pm$0.3} & 0.4$\pm$0.1 \\
                         & 120k & 0.6 & 2.5 & 0.6 & \textbf{7.0} & 1.2 \\
    \bottomrule
  \end{tabular}
\end{table}
\fi
\ifanonymous\else\section*{Acknowledgment}
Large language models, specifically Anthropic's Claude models, were used as coding
assistants and to assist with writing and editing the manuscript. The authors reviewed
all generated code and text and take full responsibility for the content of this paper.
\fi

\bibliographystyle{IEEEtran}
\bibliography{main}

\end{document}